\documentclass[a4paper]{article}
\usepackage[english]{babel}
\usepackage[utf8]{inputenc}
\usepackage[T1]{fontenc}
\usepackage[a4paper,top=3cm,bottom=2cm,left=3cm,right=3cm,marginparwidth=1.75cm]{geometry}
\usepackage[colorinlistoftodos]{todonotes}
\usepackage[colorlinks=true, allcolors=blue]{hyperref}
\graphicspath{{figures/}}
\usepackage{caption}
\usepackage{subcaption}
\usepackage{authblk}
\usepackage{float}     

\usepackage{url}

\usepackage{amsmath,amsfonts,amssymb}
\usepackage{graphics,graphicx}
\usepackage{color}
\usepackage[misc]{ifsym}  % for envelope icon
\usepackage{booktabs}
\usepackage{multirow}

\newcommand{\keywords}[1]{\bigskip \noindent {\bf Keywords:} \ #1}

\title{Topological descriptors of foot clearance gait dynamics improve differential diagnosis of Parkinsonism}

\author{Jhonathan Barrios$^{1}$$^{*}$, Wolfram Erlhagen$^{1}$, Miguel F. Gago$^{2,3}$, Estela Bicho$^{4}$ and Flora Ferreira$^{5,6}$ \\
        \small $^{1}$Centre of Mathematics (CMAT), School of Sciences,University of Minho, Portugal \\
        \small $^{2}$Neurology Departmet, Hospital da Senhora da Oliveira, Portugal \\
        \small $^{3}$ ICVS, School of Medicine, University of Minho, Portugal\\
        \small $^{4}$ Algoritmi Centre, School of Engineering, University of Minho, Portugal\\
         \small $^{5}$ Centre of Mathematics (CMUP), University of Porto, Portugal\\
          \small $^{6}$ School and Economics and Management, University of Porto, Portugal\\
        \small $^{*}$Corresponding author: \tt{jhonathanbarrios21@gmail.com} \\
}
\date{}

\begin{document}
\maketitle

%%%% Abstract %%%%
\begin{abstract} 

Differential diagnosis among parkinsonian syndromes remains a clinical challenge due to overlapping motor symptoms and subtle gait abnormalities. Accurate differentiation is crucial for treatment planning and prognosis. While gait analysis is a well-established approach for assessing motor impairments, conventional methods often overlook hidden nonlinear and structural features embedded in foot clearance patterns. We evaluated Topological Data Analysis (TDA) as a complementary tool for Parkinsonism classification using foot-clearance time series.  Persistent homology produced Betti curves, persistence landscapes, and silhouettes, which were used as features for a Random Forest classifier. 
The dataset comprised 15 controls (CO), 15 idiopathic Parkinson's disease (IPD), and 14 vascular Parkinsonism (VaP). Models were assessed with leave-one-out cross-validation (LOOCV). Betti-curve descriptors consistently yielded the strongest results. 
For IPD vs VaP, foot-clearance variables-minimum toe clearance, maximum toe late swing, and maximum heel clearance—achieved 83\% accuracy and AUC=0.89 under LOOCV in the medicated (On) state. Performance improved in the On state and further when both Off and On states were considered, indicating sensitivity of the topological features to levodopa related gait changes. These findings support integrating TDA with machine learning to improve clinical gait analysis and aid differential diagnosis across parkinsonian disorders.
\end{abstract}
\noindent\keywords{topological data analysis; persistent homology; differential diagnosis; foot clearance pattern; Parkinsonism}

%%%% Section1 %%%%
\section{Introduction}
Gait analysis has gained important recognition as a valuable clinical tool, providing an objective evaluation of gait impairment \cite{phinyomark2018analysis}. It plays a crucial role in monitoring disease progression and evaluating the effectiveness of therapeutic interventions \cite{fernandes2021}. In parallel, time series data analysis has advanced significantly with the development of new techniques capable of extracting valuable predictive information from complex, high-dimensional structures \cite{fernandes2021,yan2020classification}. In this context, human gait dynamics represents an excellent example of a complex process characterized by nonlinear relationships between multiple inputs and outputs \cite{fernandes2021,yan2020classification}.

Deviations from normal gait patterns, defined as gait disturbances, can be the result of compensatory mechanisms where the body adapts to maintain mobility, sometimes leading to suboptimal movement patterns \cite{Ataullah2024}. Chronic gait disturbances, particularly those associated with neurological dysfunctions, have a profound impact on patients’ quality of life, morbidity, and mortality. Given their clinical significance, gait abnormalities have been extensively studied as potential biomarkers of neurological disorders, including Parkinson’s disease and related conditions \cite{Ataullah2024}.

In the context of Parkinsonism, the differential diagnosis between idiopathic Parkinson’s disease (IPD) and vascular Parkinsonism (VaP) is crucial for clinical management and treatment decisions. However, the phenotypic similarities between IPD and VaP put significant challenges in accurately distinguishing these two parkinsonian disorders \cite{Zijlmans2004,lehosit2015,schaaf2017}. In addition, the effect of medication (levodopa) on gait variability remains inconclusive. Some studies have reported that levodopa reduces the variability of specific gait parameters such as step time, swing time, stride length, and stride velocity \cite{Bryant2011,Bryant2016}. In contrast, other findings do not indicate a significant impact on the variability of double support time and postural control \cite{Bryant2016,Galna2015}. This distinction is particularly relevant in the context of VaP, where cerebrovascular lesions can unevenly affect both dopaminergic and nondopaminergic pathways. Moreover, studies of gait analysis that focus on VaP are limited \cite{Zijlmans2004}, and to date, only this work \cite{ferreira2019} has specifically examined gait variability and foot clearance patterns.

Previous studies have explored gait analysis as a promising tool for capture subtle motor differences between these conditions. Among these efforts, normalization techniques based on multiple linear regression (MLR) models have demonstrated the potential to mitigate the influence of individual physical characteristics on gait measurements, thus improving the accuracy of classification \cite{fernandes2021,ferreirabarrios2023,barrios2025}. Although linear approaches like MLR have shown promise in reducing variability and improving gait normalization, they are inherently limited in their ability to capture the complex, nonlinear dynamics present in gait data. To address these limitations, advanced analytical techniques are needed to better characterize these intricate patterns and uncover subtle differences in motor function. In turn, gait analysis of stride-to-stride variability and foot clearance patterns helps objectively capture the clinical changes observed in VaP, such as foot dragging and extended posture \cite{ferreira2019}. Increased gait variability in VaP may be associated with a reduced stride length, cognitive impairment, and damage to the nondopaminergic brain regions that are less responsive to levodopa \cite{ferreira2019}. These findings highlight the need for further studies to better understand the VaP phenotype in contrast to IPD.

Topological Data Analysis (TDA) has emerged as a powerful framework for analyzing complex and nonlinear systems, offering unique insights by examining the geometric and topological structures underlying time series data \cite{karan2021}. In the context of gait analysis, TDA provides a novel approach to capture the dynamic interplay of spatio-temporal parameters, enabling a more robust discrimination between neurodegenerative conditions such as Parkinson’s disease \cite{yan2020classification,yan2022topological,Barrios2025AMMS}. For example, a topological motion analysis framework has been proposed to classify gait fluctuations by embedding time series into phase space and extracting persistent homology features from barcodes \cite{yan2020classification}. When combined with machine learning classifiers, this approach successfully distinguished gait patterns associated with diseases such as amyotrophic lateral sclerosis, Huntington’s disease, and Parkinson’s disease from those of healthy controls \cite{yan2020classification}. Furthermore, TDA has been applied to detect freezing-of-gait episodes in Parkinson’s patients by analyzing reconstructed phase space data using topological descriptors such as Betti Curves (BC), Persistence Landscapes (PL), and Silhouette Landscapes (SL). These features outperformed traditional complexity metrics, demonstrating the potential of TDA to advanced gait analysis in neurodegenerative disorders \cite{yan2022topological,Barrios2025AMMS}.

This work aimed to explore the potential of TDA, specifically persistent homology, to characterize gait dynamics in Parkinson’s disease using gait foot clearance variables, underused variables in the literature. Although previous studies have highlighted the promise of TDA in gait analysis, its application remains limited in scope. Notably, foot clearance measures have been underused in this context and, to the best of our knowledge, have not yet been paired with TDA for differential diagnosis. We first computed persistence diagrams from foot clearance time series, then derived three topological descriptors (BC, PL and SL) to summarize nonlinear structure.  These descriptors were combined with a Random Forest classifier to: (i) discriminate between control subjects (CO) and IPD, and between CO and VaP, and (ii) distinguish IPD from VaP, while also evaluating how levodopa (unmedicated (Off) vs medicated (On)) modulates classification performance. This integrated topology aware machine learning framework broadens prior work by targeting clinically relevant differentiation of parkinsonian subtypes.

This work is organized as follows: Section 2 reviews related works on the application of TDA, emphasizing the use of algebraic topology tools for classification tasks in gait time series analysis. Section 3 describes the gait dataset and the proposed pipeline to extract topological descriptors and integrate TDA with machine learning algorithms. Section 4 presents the results, highlighting the classification performance of topological descriptors across different gait variables and medication states. Section 5 provides a detailed discussion of these findings, addressing their methodological relevance, clinical implications, and potential to inform future research in the analysis of gait dynamics in parkinsonian disorders. Finally, Section 6 provides conclusions and outlines future directions for this research.

\section{Persistent Homology and Topological Descriptors}
In general, TDA can be described as a collection of methods for identifying and analyzing structure within data. These methods encompass techniques such as clustering, manifold estimation, nonlinear dimensionality reduction, mode estimation, ridge estimation, and persistent homology \cite{wasserman2018,chazal2021}. In this section, the TDA concepts relevant to this study are formally introduced, with a particular focus on persistent homology and topological descriptors applied to the work.

\subsection{Persistent Homology}
Homology provides a quantitative framework to describe the connectivity of a space, offering a mathematical approach to identify and analyze the presence of features such as connected components, loops, and voids within a point cloud \cite{edelsbrunner2010}. Persistent homology captures multiscale topological features by analyzing a nested sequence of simplicial complexes built from data \cite{chazal2021}. This approach tracks how homological features emerge, persist and eventually disappear on different scales, providing a robust method for encoding the topology of the data.

In practical applications, persistent homology is often computed on point clouds derived from data using methods such as time-delay embedding. From these point clouds, a filtration is constructed by systematically expanding the scale parameter \( r \), resulting in a sequence of nested complexes. The topological features are then recorded as they appear and disappear during this process, creating intervals that form the basis for persistence diagrams or barcodes \cite{chazal2021}. 

To understand this process, consider a point cloud \( C \) where a filtration is generated by the union of expanding balls centered on the points in \( C \) \cite{chazal2021}. The evolution of the topological features in this filtration can be summarized as follows:

\begin{itemize}
    \item At \( r = 0 \), the union of balls corresponds to the original point cloud, where each point represents a zero-dimensional feature (a connected component). An interval is initiated for the birth of each feature.
    \item As \( r \) increases, the balls begin to overlap, causing some connected components to merge. The persistence diagram records the death of these features by closing their corresponding intervals.
    \item Further expansion leads to the formation of higher-dimensional features, such as loops. These features were born at specific scales, represented by new intervals in the persistence diagram.
    \item As the scale continues to grow, some loops and voids are filled in, signaling their death. The corresponding intervals in the barcode are closed.
    \item At the largest scales, all features have either merged or disappeared, leaving only the most persistent features, which provide insights into the overall topology of the data.
\end{itemize}

The persistence barcode encodes the birth and death of features on all scales. The length of each interval reflects the persistence of its associated feature, with longer intervals representing more significant topological structures \cite{chazal2021}. These barcodes can be transformed into persistence diagrams, where each interval \((a, b)\) is represented as a point \((a, b)\) in \( \mathbb{R}^2 \) \cite{chazal2021}.

\subsection{Topological Descriptors}

BC, PL, and SL serve as powerful topological descriptors derived from the nonlinear dynamics of gait time series \cite{yan2020classification}. These descriptors encode intricate properties of the gait dynamics by capturing the structure and persistence of topological features in the phase space. 

\subsubsection{Betti Curves (BC)}
Betti curves are a simplified representation derived from persistent homology that describes how topological features, such as connected components, loops, or cavities, evolve across scales in a filtration \cite{chazal2021}. These features are captured by the Betti numbers (\(\beta_k\)), where \(\beta_k\) denotes the number of independent \(k\)-dimensional topological features \cite{wasserman2018}, specifically: \(\beta_0\) represents the number of connected components,  \(\beta_1\) indicates the number of loops (1 dimensional holes), and  \(\beta_2\) corresponds to the number of cavities (2-dimensional holes).

In practice, for a persistence diagram composed of triples of the birth-death dimension \([b, d, q]\), subdiagrams associated with different homology dimensions are individually analyzed. The Betti curves for each dimension are then derived by uniformly sampling the filtration parameter \cite{Guzel2023}.

To formalize this, let \(\text{dgm} = \{(b_i, d_i) \mid i \in I\}\) represent a persistence diagram. The Betti curve corresponding to \(\text{dgm}\) is defined as the function \( \beta_{\text{dgm}} : \mathbb{R} \to \mathbb{N} \), where the value at \( t \in \mathbb{R} \) indicates the number of points \((b_i, d_i)\) in \(\text{dgm}\) for which \( b_i \leq t < d_i\). Formally, this can be expressed as:
\[
\beta_{\text{dgm}}(t) = \#\{(b_i, d_i) \in \text{dgm} \mid b_i \leq t < d_i\}.
\]

Betti curves can be viewed as a projection of the persistence diagram, summarizing the evolution of topological features across the filtration scale. They provide a compact, yet informative representation that focuses on the count of features in a specific dimension \(k\) as a function of the filtration parameter \(t\) \cite{wasserman2018}. 

\subsubsection{Persistence Landscapes (PL)}
The persistence landscape offers an alternative method to represent persistence diagrams. This method aims to translate the topological information captured in persistence diagrams into elements within a Hilbert space, allowing the direct application of statistical learning techniques \cite{bubenik2015,BUBENIK2017}. The persistence landscape is defined as a set of continuous piecewise linear functions \(\lambda: \mathbb{N} \times \mathbb{R} \rightarrow \mathbb{R}\) that effectively summarize a persistence diagram \((\text{dgm})\) \cite{bubenik2015,BUBENIK2017,Chazal2014}.

Suppose a birth-death pair \(p = (b, d) \in \text{dgm}\) is transformed into to the point \(\left(\frac{b+d}{2}, \frac{d-b}{2}\right)\). Note that points with infinite persistence are excluded in this context. The landscape is constructed by creating functions that tent over the features of the rotated persistence diagram, as follows:

\begin{equation}
    \Lambda_p(t) =
\begin{cases} t-b, & \text{if } t \in \left[b, \frac{b+d}{2}\right], \\
d-t, & \text{if } t \in \left(\frac{b+d}{2}, d\right], \\
0, & \text{otherwise}.
\end{cases}
\end{equation}

The persistence landscape \(\lambda_{\text{dgm}}\) associated with \(\text{dgm}\) provides a summary of the piecewise linear curves by overlaying the graphs of the functions \(\{\Lambda_p\}_{p \in \text{dgm}}\) \cite{Chazal2014}.

\subsubsection{Silhouettes Landscapes (SL)}
Silhouette landscapes offer a weighted summary of a persistence diagram by averaging the triangular functions $\Lambda_p(t)$, using persistence-based weights \cite{Chazal2014}. For a persistence diagram consisting of \(m\) off-diagonal points, the weighted silhouette function \(\varphi(t)\) is defined as the weighted average of triangular functions. Specifically, it can be expressed as follows:

\begin{equation}
    \varphi(t) = \frac{\sum_{j=1}^{m} w_j \Lambda_j(t)}{\sum_{j=1}^{m} w_j},
\end{equation}

where the weights \(w_j\) are chosen based on the persistence of each point in the diagram. Given two points in the persistence diagram, represented by the pairs \((b_i, d_i)\) and \((b_j, d_j)\), the weights satisfy the condition \(w_j \geq w_i\) whenever \(|d_j - b_j| \geq |d_i - b_i|\). In this case, the weights are defined as \(w_j = |d_j - b_j|^p\), where \(p > 0\) \cite{Chazal2014}.

\noindent The power-weighted silhouette is defined as follows:

\begin{equation}
    \varphi^{(p)}(t) = \frac{\sum_{j=1}^{m} |d_j - b_j|^p \Lambda_j(t)}{\sum_{j=1}^{m} |d_j - b_j|^p},
\end{equation}

the parameter \(p\) serves as a trade-off between giving equal importance to all pairs in the persistence diagram and focusing on the most persistent. For small values of \(p\), the silhouette is influenced more by low persistence pairs, whereas for large \(p\), it is dominated by the most persistent pairs \cite{Chazal2014}.

These descriptors collectively offer a comprehensive view of the dynamic topological behavior of gait, serving as a solid foundation for classification tasks.

%%%%
\section{Methods}
\subsection{Gait Data}
The study population was recruited from the Department of Neurology of Senhora da Oliveira Hospital, in the municipality of Guimarães, Portugal. The ethics committee of the local hospital approved the study protocol, which was submitted by the Life and Health Sciences Research Institute of the University of Minho (UM) and the Algoritmi/UM Center. Written consent was obtained from all subjects or their guardians.

Patients with VaP and IPD were consecutively selected from outpatient consultations on movement disorder. The diagnosis was based on published clinical criteria \cite{Zijlmans2004} and was supported by a retrospective clinical history, with a longitudinal reassessment of the clinical diagnosis. The exclusion criteria for all patients included the presence of resting tremor, moderate to severe dementia (CDR $>$ 2), musculoskeletal diseases, and evident clinical progression since diagnosis (Hoehn-Yahr $>$ 3). 

Table \ref{table:physicalCha} summarizes the anthropometric data of the study groups related to Parkinsonism and the control group. At the end of the recruitment phase, gait data were collected from 14 VaP patients and 15 IPD patients (Table \ref{table:physicalCha}). In addition, 15 healthy adults (CO group) of different ages and sexes were recruited for gait data collection to to support differentiation between CO and Parkinsonism. 

\begin{table}[h]
\caption{Physical Characteristics for the study groups}
\begin{center}
\begin{tabular}{lcccc}
\hline
      & Controls & IPD Patients & VaP Patients \\
\hline
Age (years) & $76.00 \pm 5.70$ & $76.60 \pm 4.29$ & $80.53 \pm 4.63$ \\
Weight (kg) & $68.21 \pm 7.03$ & $73.24 \pm 12.53$ & $66.17 \pm 10.38$ \\
Height (m) & $1.63 \pm 0.06$ & $1.67 \pm 0.082$ & $1.61 \pm 0.085$  \\
Gender (Female/Male) & 9/6 & 4/11 & 6/8 \\
\hline
\end{tabular}
\smallskip
Note: Data are displayed as mean $\pm$ standard deviation
\end{center}
\label{table:physicalCha}
\end{table}

Gait data were collected using two Physilog® sensors (Gait Up®, Switzerland) attached to the dorsum of each shoe with elastic bands. Participants walked a 60-meter continuous course, consisting of a 30-meter corridor with one turn, at a self-selected speed. Patients were evaluated in OFF phase (24h without levodopa) and in ON phase (after a supra-threshold levodopa challenge, 150\% of L-dopa morning dose).

\begin{figure}[!ht]
    \begin{center}
        \includegraphics[scale=0.23]{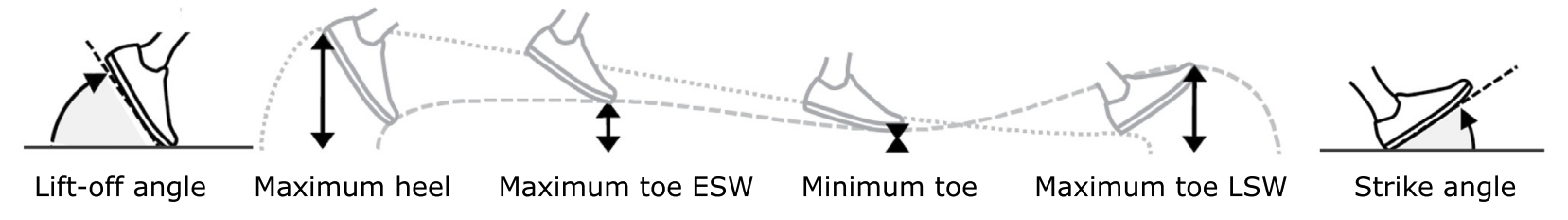}
    \end{center}
\caption{Foot clearance variables (Adapted from Gait Up® document support)}
\label{fig:FootClearance}
\end{figure}

\noindent Using Gait Up® software, which incorporates a proprietary fusion algorithm based on gait event detection, signal drift correction, strap integration, and biomechanical modeling, this study evaluated a specific group of variables related to foot clearance (see Figure \ref{fig:FootClearance}), including Lift-off Angle, maximum heel clearance (MaxHC), maximum toe early swing (MaxTESW), minimum toe clearance (MinTC), maximum toe late swing (MaxTLSW), and Strike Angle.

This cohort has been analyzed in prior studies focusing on conventional gait features \cite{fernandes2021,ferreira2019}; here we investigate topology derived foot clearance descriptors. 

\subsection{Methodology}

\begin{figure}[!ht]
    \begin{center}
        \includegraphics[scale=0.65]{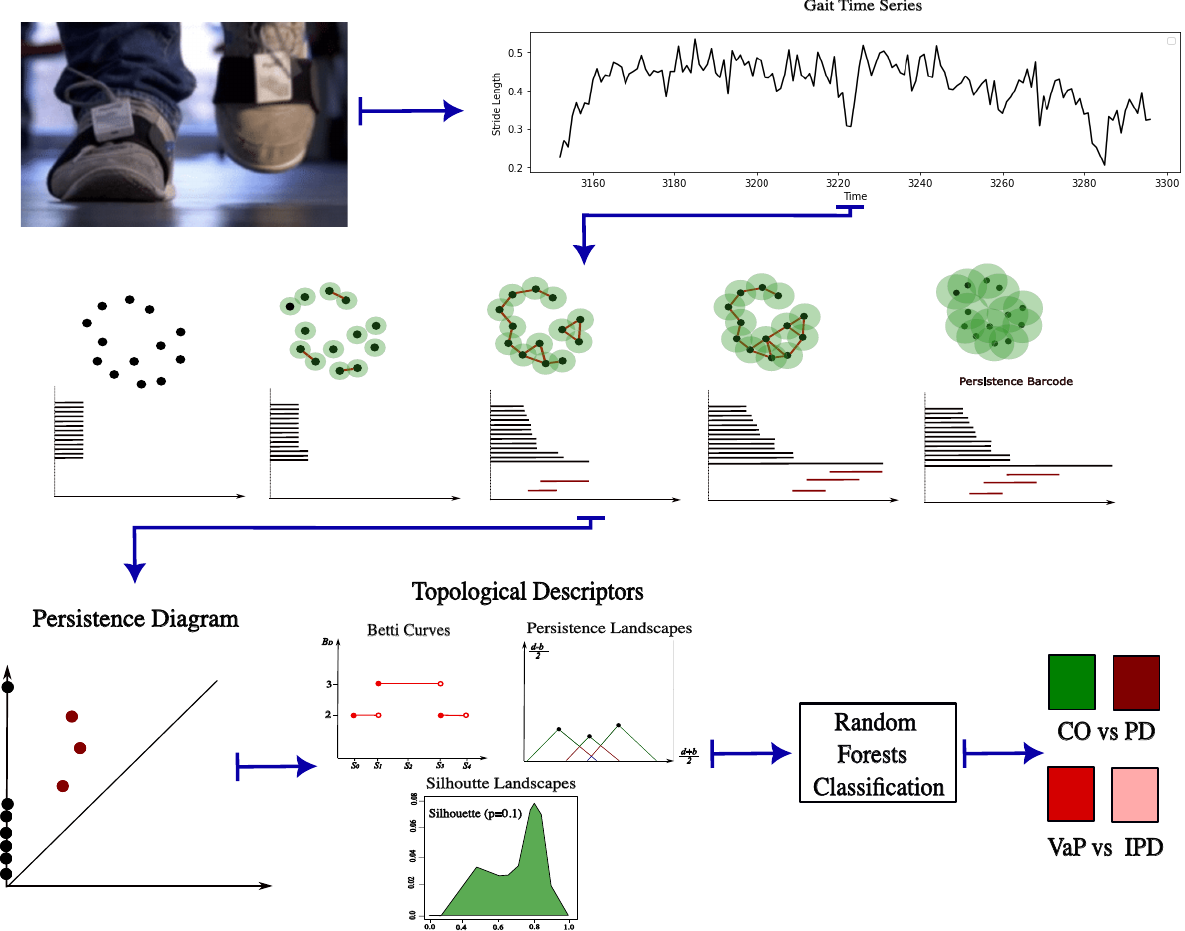}
    \end{center}
\caption{The proposed complementary framework for differential Parkinsonism classification based on topological descriptors and Random Forest}
\label{fig:TDApipeline}
\end{figure}

Figure \ref{fig:TDApipeline} displays the methodology used to integrate topological descriptors into the classification task followed a procedure analogous to that reported in a previous study \cite{Barrios2025AMMS}. The process began with the phase space reconstruction of the time series corresponding to the variable under study. To achieve this, the time series was embedded using a time-delay embedding technique, transforming the data into a point cloud representation. Specifically, the \textit{TakensEmbedding} transformer from the \texttt{gtda.time\_series} module of the \texttt{giotto-tda} library in Python was used. \textit{Giotto-TDA} is a high performance topological machine learning toolbox designed for efficient computation of topological features \cite{tauzin2020giotto}. This embedding process allows the one-dimensional time series to be mapped to a higher-dimensional space while preserving key dynamic properties of the original data. In this study, an embedding dimension of \( d = 2 \) was used, which has been shown to effectively capture the essential features of gait dynamics.

Following the embedding step, persistent homology computation was performed on the generated point clouds using the Vietoris–Rips complex \cite{Chazal2013VietorisRips}. First, persistence diagrams were constructed to track the birth and death of topological features, such as connected components and loops, on multiple scales. In this paper, homology dimensions \(H_0\) and \(H_1\) were used to provide information on different topological features: \(H_0\) corresponds to connected components, reflecting the emergence and disappearance of distinct clusters, while \(H_1\) captures loops, indicating cyclic patterns in time series data. The \texttt{Ripser} package from the \texttt{scikit-tda} library in Python was used to compute persistence diagrams \cite{scikittda2019}. Next, other representations of persistent homology, the topological descriptors BC, PL and SL were extracted for each persistence diagram. Descriptors were computed on a fixed grid of $nbins=25$ points per homology degree; concatenating $H_0$ and $H_1$ yields 50 values per descriptor. We used $(d,\tau,nbins)=(2,1,25)$ across experiments to balance resolution and model complexity on this small dataset.

In the feature aggregation step, the topological features extracted  from BC, PL, and SL were calculated for each gait variable.  These features were then concatenated to form the final feature vectors for each subject, which were stored in structured arrays for further analysis.

Finally, a Random Forest was chosen for classification due to its versatility and proven effectiveness in previous works \cite{fernandes2021,barrios2025}. The extracted vectors from these descriptors were then used as features to train the RF model in each classification task, where, to ensure robust model evaluation,subject wise leave-one-out cross-validation (LOOCV) was applied, holding out one subject per fold for testing.  Model performance was assessed using multiple metrics, including accuracy, area under the receiver operating characteristic curve (AUC), sensitivity, specificity, and confusion matrices. We analyzed for CO vs IPD, CO vs VaP, and for IPD vs VaP, and we further assessed the effect of levodopa by evaluating in the Off and On states separately as well as jointly (Off+On).

This process provided a comprehensive method for analyzing and classifying gait time series data using topological descriptors, allowing us to capture global and subtle variations in the dynamics of gait data, resulting in improved classification performance.

\section{Results}
Classification results are presented for distinguishing controls from parkinsonian subtypes (CO vs IPD, CO vs VaP) and for discriminating idiopathic and vascular Parkinsonism (IPD vs VaP) using subject wise LOOCV in the Off, On, and combined Off+On conditions. Performance is summarized by AUC, accuracy (Accu), sensitivity (Sens), and specificity (Spec).

\subsection{Distinguishing controls from parkinsonian subtypes}
Table \ref{tab:ResultsCOvsPD} presents results for CO vs IPD and CO vs VaP using three topological descriptors (BC, PL, and SL) applied to the foot clearance variables. BC consistently ranked first across variables and tasks, PL showed intermediate performance, SL performed worst.

BC delivered near perfect discrimination for most variables in both CO vs IPD and CO vs VaP (AUC 0.99–1.00; Acc/Sens/Spec $\ge$ 0.93). In CO vs IPD, BC reached AUC=1.00 for five of six variables, with perfect accuracy for Lift-off Angle, MaxHC, MaxTLSW, and Strike Angle. Similarly, in CO vs VaP, BC achieved AUC=1.00 for all variables and perfect results for Strike Angle, indicating robustness despite the greater clinical heterogeneity in VaP. PL was clearly inferior to BC but acceptable results, particularly for the MinTC and MaxTESW variables in CO vs VaP. However, both the specificity and sensitivity of PL were more variable, and performance dropped on MaxTLSW. In CO vs IPD, PL performed best on Lift-off Angle and MaxTESW (AUCs of 0.85 and 0.84), but with a marked decrease in other variables. In CO vs VaP, PL managed to remain above 0.85 for most variables, although it showed notable drops in MaxTLSW (AUC = 0.91, accuracy = 75\%, sensitivity = 57\%), suggesting some vulnerability in contexts of greater clinical complexity such as VaP. Finally, SL performed worst across tasks, with low AUC, sensitivity, and specificity values, especially for variables such as MaxTLSW and Strike Angle. In CO vs IPD, SL barely achieved AUCs above 0.6 for most variables, and in MaxTESW it dropped to 0.40. In CO vs VaP, performance was even lower: in MaxTLSW, SL obtained an AUC of only 0.11 and a sensitivity of 7\%, suggesting that this descriptor is highly sensitive to noise or heterogeneity.

\begin{table}[t]
\centering
\caption{Classification performance metrics obtained based in the topological descriptor vectors for distinguishing controls from parkinsonian subtypes}
{\footnotesize %\small
\begin{tabular}{llcccccccc}
\toprule
  & & \multicolumn{4}{c}{CO vs VaP} & \multicolumn{4}{c}{CO vs IPD} \\ 
\cmidrule(lr){3-6} \cmidrule(lr){7-10} 
Gait Variable  & TD & AUC & Accu &  Sens & Spec & AUC & Accu &  Sens & Spec  \\
\midrule
 Lift-off Angle & {\bf BC} & {\bf 1.00} & {\bf 1.00} & {\bf 1.00} & {\bf 1.00} & {\bf 1.00} & {\bf 1.00} & {\bf 1.00} & {\bf 1.00} \\
                & PL & 0.89 & 0.82 & 0.71 & 0.93 & 0.85 & 0.76 & 0.67 & 0.87 \\
                & SL & 0.53 & 0.58 & 0.64 & 0.53 & 0.62 & 0.53 & 0.53 & 0.53 \\
\midrule
MaxHC           & {\bf BC} & {\bf 1.00} & {\bf 0.97} & {\bf 0.93} & {\bf 1.00} & {\bf 1.00} & {\bf 1.00} & {\bf 1.00} & {\bf 1.00} \\
                & PL & 0.85 & 0.72 & 0.64 & 0.80 & 0.82 & 0.73 & 0.73 & 0.73 \\
                & SL & 0.71 & 0.68 & 0.64 & 0.73 & 0.65 & 0.70 & 0.67 & 0.73 \\
\midrule
MaxTESW          & {\bf BC} & {\bf 1.00} & {\bf 0.93} & {\bf 0.86} & {\bf 1.00} & {\bf 0.99} & {\bf 0.97} & {\bf 1.00} & {\bf 0.93} \\
                & PL & 0.97 & 0.89 & 0.79 & 1.00 & 0.84 & 0.76 & 0.73 & 0.80 \\
                & SL & 0.79 & 0.79 & 0.64 & 0.93 & 0.40 & 0.43 & 0.33 & 0.53 \\
\midrule
MinTC           & {\bf BC} & {\bf 1.00} & {\bf 0.96} & {\bf 0.93} & {\bf 1.00} & {\bf 0.99} & {\bf 0.93} & {\bf 0.93} & {\bf  0.93} \\
                & PL & 0.98 & 0.93 & 0.93 & 0.93 & 0.87 & 0.80 & 0.60 & 1.00 \\
                & SL & 0.65 & 0.62 & 0.71 & 0.53 & 0.61 & 0.63 & 0.60 & 0.67 \\
\midrule
MaxTLSW          & {\bf BC} & {\bf 1.00} & {\bf 1.00} & {\bf 1.00} & {\bf 1.00} & {\bf 1.00} & {\bf 1.00} & {\bf 1.00} & {\bf 1.00} \\
                & PL & 0.91 & 0.75 & 0.57 & 0.93 & 0.78 & 0.70 & 0.67 & 0.73 \\
                & SL & 0.11 & 0.27 & 0.07 & 0.47 & 0.75 & 0.66 & 0.67 & 0.67 \\
\midrule
Strike Angle    & {\bf BC} & {\bf 1.00} & {\bf 1.00} & {\bf 1.00} & {\bf 1.00} & {\bf 1.00} & {\bf 0.97} & {\bf 0.93} & {\bf 1.00} \\ 
                & PL & 0.88 & 0.79 & 0.79 & 0.80 & 0.78 & 0.70 & 0.67 & 0.73 \\
                & SL & 0.40 & 0.51 & 0.43 & 0.60 & 0.72 & 0.70 & 0.73 & 0.67 \\
\bottomrule
\end{tabular}
}
\smallskip
  \begin{flushleft}
    \scriptsize{TD: Topological Descriptor; Accu: Accuracy; Sens: Sensitivity; Spec: Specificity ; MaxHC: Maximum Heel Clearance;  MaxTESW: Maximum Toe Early Swing; MinTC: Minimum Toe Clearance; MaxTLSW: Maximum Toe Late Swing.}
  \end{flushleft}
\label{tab:ResultsCOvsPD}
\end{table}

Despite the greater clinical heterogeneity in VaP, BC maintained high performance, indicating robustness and generalizability. In contrast, PL showed relatively better performance in CO vs VaP than in CO vs IPD for certain variables, such as MinTC, which could be related to the way topological features are distributed in patients with vascular damage. SL, on the other hand, showed low levels of specificity in both tasks, suggesting a higher number of false positives. This limitation could be due to its sensitivity to small perturbations in the persistence diagrams, making it less reliable in contexts with small datasets as demonstrated in \cite{Barrios2025AMMS}. The results obtained in this study reinforce the ability of Betti Curves to capture essential aspects of the underlying gait dynamics that are not fully represented by traditional statistical metrics. Betti Curves directly describe the evolution of connected components \(H_0\) and loops \(H_1\) through different filtering thresholds, providing a compact and stable representation of the overall topological structure of the system. This property is especially relevant for signals with low temporal resolution, such as the gait series analyzed in this study, where each signal is composed of approximately 36 observations per gait cycle. In this type of data, phase space reconstruction is limited, and more global descriptors like BC tend to better preserve dominant structural information without amplifying noise. In contrast, in the previous study based on a different type of dataset \cite{Barrios2025AMMS}, the signals exhibit higher temporal resolution (101 observations per cycle) and lower relative variability between subjects. In this context, Persistence Landscapes demonstrated superior performance, likely because their functional and multiscale nature allows them to exploit the greater density of temporal information to identify subtle variations in signal morphology. However, this same sensitivity can become unfavorable in shorter or heterogeneous series, where local fluctuations can be mistaken for noise or individual variability.

This comparison suggests that the choice of topological descriptor should be guided by the nature and complexity of the dataset. While PL is advantageous in domains with high resolution and smooth signals, BC offers a more robust and stable alternative in clinical settings with limitations in sample size or recording quality. These observations are consistent with the findings of \cite{yan2020classification,yan2022topological,chazal2021,bubenik2015}, which highlight that the stability of topological descriptors strongly depends  on the level of discretization, the dimensionality of the embedded space, and the presence of noise in the data.

\subsection{Discriminating Idiopathic and Vascular Parkinsonism}
Based on previous results obtained in the classification of CO from PD subtypes, the BC topological descriptor vector was selected to address this task. In addition, the effect of levodopa on the discriminatory capacity of the model was evaluated separately, considering the Off state (without medication), the On state (after levodopa administration), and a combined condition (Off+On) that incorporates data from both states.

The results in Table \ref{tab:ResultsIPDvsVaP} show that the performance of the model is significantly lower in the Off state, where the AUC values do not exceed 0.72 for any variable, and accuracies range between 48\% and 69\%. In particular, variables such as Lift-off Angle (AUC = 0.58, accuracy = 48\%) and MaxHC (AUC = 0.57, accuracy = 55\%) show low discriminatory ability in the absence of medication. This suggests that, in the Off state, the topological differences between IPD and VaP are less pronounced or more heterogeneous, complicating the classification based on gait dynamics.

\begin{table}[H]
\centering
\caption{Classification performance metrics obtained based on the topological descriptor vectors for IPD vs VaP}
{\small
\begin{tabular}{llcccc}
\toprule
  & & \multicolumn{4}{c}{Metric} \\ 
\cmidrule(lr){3-6}
Gait Variable  & State & AUC & Accuracy &  Sensitivity & Specificity  \\
\midrule
 Lift-off Angle & Off & 0.58 & 0.48 & 0.60 & 0.36 \\
            & On & 0.75 & 0.76 & 0.80 & 0.71 \\
            & Off+On & 0.62 & 0.62 & 0.67 & 0.57 \\
\midrule
MaxHC       & Off & 0.57 & 0.55 & 0.47 & 0.64\\
            & On  & 0.67 & 0.62 & 0.80 & 0.43\\
            & Off+On & 0.60 & 0.55 & 0.60 & 0.50\\
\midrule
MaxTESW      & Off & 0.59 & 0.66 & 0.73 & 0.57\\
            & On  & 0.65 & 0.72 & 0.73 & 0.71 \\
            & Off+On & 0.63 & 0.66 & 0.73 & 0.57\\
\midrule
{\bf MinTC}       & Off & {\bf 0.71} & {\bf 0.66} & {\bf 0.67} & {\bf 0.64} \\
            & On  & {\bf 0.81} & {\bf 0.72} & {\bf 0.73} & {\bf 0.71} \\
            & Off+On & {\bf 0.82} & {\bf 0.72} & {\bf 0.80} & {\bf 0.74} \\
\midrule
{\bf MaxTLSW}      & Off & {\bf 0.72} & {\bf 0.66} & {\bf 0.60} & {\bf 0.71}\\
            & On  & {\bf 0.83} & {\bf 0.72} & {\bf 0.73} & {\bf 0.71}\\
            & Off+On & {\bf 0.86} & {\bf 0.79} & {\bf 0.80} & {\bf 0.79}\\
\midrule
Strike Angle & Off & 0.65 & 0.69 & 0.67 & 0.71\\
             & On  & 0.76 & 0.76 & 0.80 & 0.71\\
             & Off+On & 0.71 & 0.69 & 0.73 & 0.64 \\
\bottomrule
\end{tabular}
}
\smallskip
  \begin{flushleft}
    \scriptsize{MaxHC: Maximum Heel Clearance; MaxTESW: Maximum Toe Early Swing; MinTC: Minimum Toe Clearance; MaxTLSW: Maximum Toe Late Swing.}
  \end{flushleft}
\label{tab:ResultsIPDvsVaP}
\end{table}
In contrast, in the On state, a generalized improvement in model performance is observed, with substantial increases in AUC, accuracy, and sensitivity. For example, Lift-off Angle AUC increased from 0.58 (Off) to 0.75 (On), and its accuracy improved from 48\% to 76\%. Similarly, MaxTLSW reached an AUC of 0.83 in the On state, reflecting a greater differentiation between IPD and VaP after levodopa administration.

\begin{figure}[ht]
    \centering
    % row 1: MinTC
    \begin{subfigure}[b]{0.32\textwidth}
        \centering
        \includegraphics[width=\textwidth]{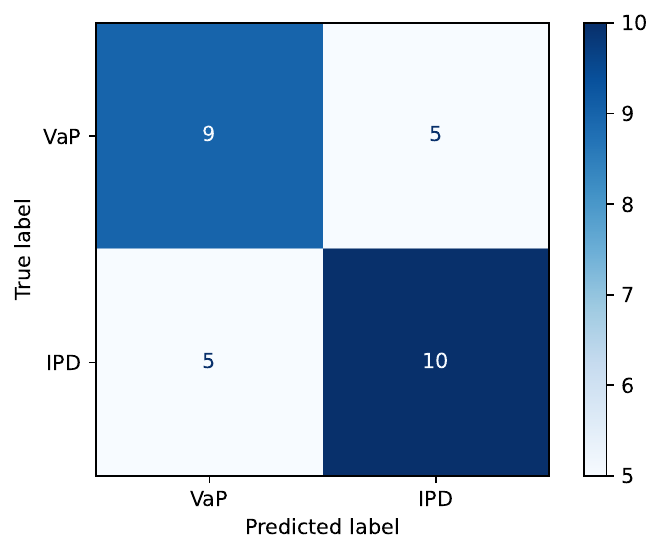}
        \caption{MinTC - Off State}
    \end{subfigure}
    \hfill
    \begin{subfigure}[b]{0.32\textwidth}
        \centering
        \includegraphics[width=\textwidth]{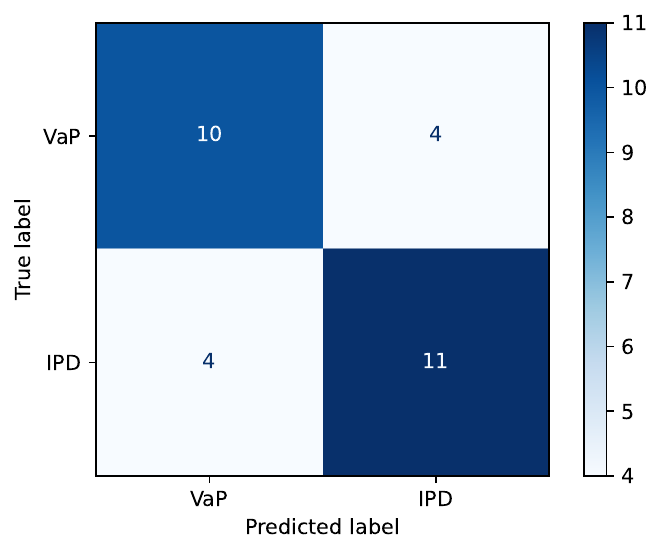}
        \caption{MinTC - On State}
    \end{subfigure}
    \hfill
    \begin{subfigure}[b]{0.32\textwidth}
        \centering
        \includegraphics[width=\textwidth]{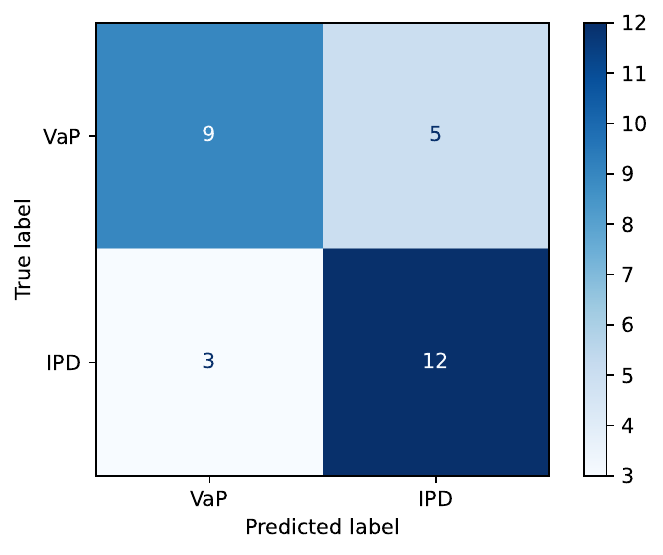}
        \caption{MinTC - Off+On}
    \end{subfigure}

    \vspace{0.5cm} % Space between rows 

    % row 2: MaxTLSW
    \begin{subfigure}[b]{0.32\textwidth}
        \centering
        \includegraphics[width=\textwidth]{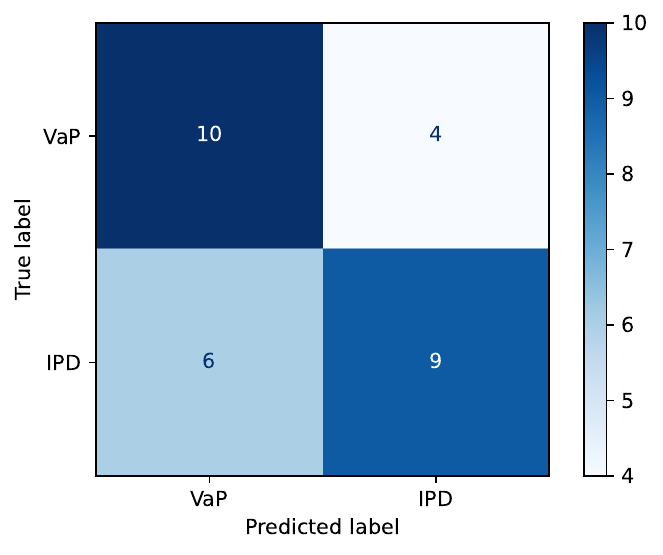}
        \caption{MaxTLSW - Off State}
    \end{subfigure}
    \hfill
    \begin{subfigure}[b]{0.32\textwidth}
        \centering
        \includegraphics[width=\textwidth]{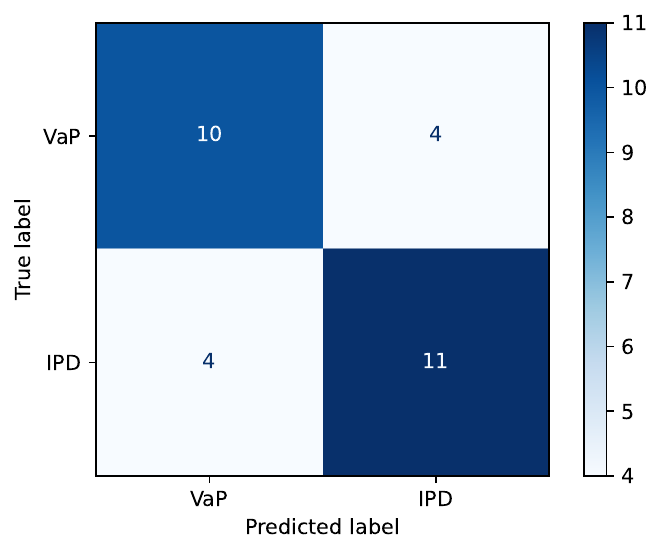}
        \caption{MaxTLSW - On State}
    \end{subfigure}
    \hfill
    \begin{subfigure}[b]{0.32\textwidth}
        \centering
        \includegraphics[width=\textwidth]{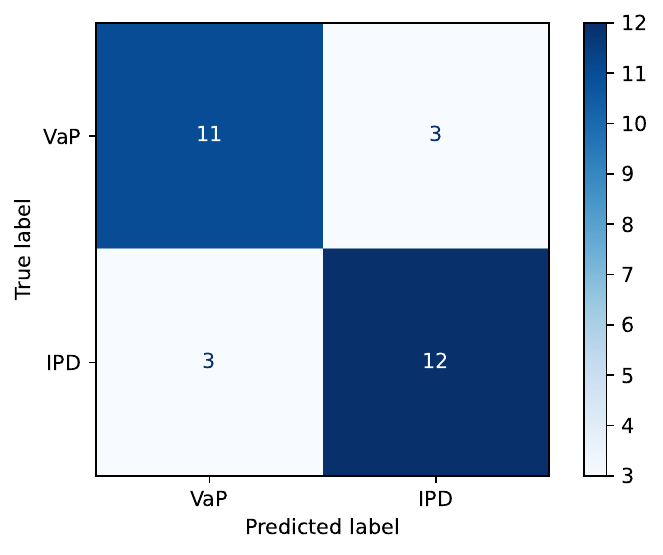}
        \caption{MaxTLSW - Off+On}
    \end{subfigure}

    \caption{Confusion matrices using the variables \textbf{MinTC} (top row) and \textbf{MaxTLSW} (bottom row) in different medication states.}
    \label{fig:ConfMatrix_MinTC_MaxTLSW}
\end{figure}

The analysis of the combined Off+On state shows performance levels that, while intermediate overall, include noteworthy improvements for specific variables. For instance, MinTC (AUC = 0.82) and MaxTLSW (AUC = 0.86) achieved higher classification performance than in either the Off or On states alone. This suggests that integrating information from both states can offer a more complete representation of gait dynamics and improve discrimination between IPD and VaP. In this sense, the temporal combination could capture the differential response patterns to medication, which adds clinical value to the topological assessment.  The confusion matrices shown in Figure \ref{fig:ConfMatrix_MinTC_MaxTLSW} provide additional insight into the classification behavior using the MinTC and MaxTLSW variables across Off, On, and Off+On medication states. These visualizations illustrate the concrete impact of levodopa on the classifier's ability to differentiate between IPD and VaP. 

In the Off state, both variables show moderate classification performance. For MinTC, 5 IPD and 5 VaP patients were misclassified (Figure \ref{fig:ConfMatrix_MinTC_MaxTLSW}a), indicating limited discriminative power without medication. Similarly, MaxTLSW misclassified 6 patients with IPD and 4 patients with VaP (Figure \ref{fig:ConfMatrix_MinTC_MaxTLSW} d), showing a slightly worse balance between sensitivity and specificity. These results reflect the known clinical overlap between IPD and VaP in the absence of dopaminergic treatment. However, in the On state, performance improved considerably for both variables. In Figure \ref{fig:ConfMatrix_MinTC_MaxTLSW}b (MinTC), the number of correctly classified IPD patients increased to 11, with only 4 misclassified VaP patients — this is consistent with the AUC of 0.81 reported for this variable. In addition, MaxTLSW in Figure \ref{fig:ConfMatrix_MinTC_MaxTLSW}e shows a similar improvement, with 11 IPD patients correctly identified and only 4 errors in VaP classification, supporting the strong AUC of 0.83 for this condition. These findings suggest that levodopa improves the differentiation between subtypes by modulating the dynamic properties of gait captured through foot clearance. Figures \ref{fig:ConfMatrix_MinTC_MaxTLSW}c and \ref{fig:ConfMatrix_MinTC_MaxTLSW}f reveal the highest classification performance. For MinTC, 12 IPD patients and 9 VaP patients were correctly classified, while MaxTLSW shows 12 correct IPD classifications and 11 correct VaP classifications—an excellent result with minimal misclassification (AUC = 0.86). This suggests that combining information from both medication states allows the model to better capture temporal variations and medication response patterns, thereby improving discrimination even in overlapping clinical presentations. 

Having examined the classification performance of each gait variable, we evaluated the possibility that combining multiple gait variables could improve the model performance. To this end, multivariate input vectors were constructed by concatenating the Betti Curves descriptors corresponding to different gait variables. For example, combining two variables consisted of concatenating the BC vectors of Variable 1 and Variable 2 into a single input vector for the classifier. This procedure was systematically applied to all possible combinations of 2, 3, 4, and 5 variables, as well as to the combination of all variables, with the aim of determining whether integrating topological information from multiple variables improves the classifier's ability to capture subtle and distributed patterns that distinguish IPD from VaP. This strategy hypothesizes that nonlinear gait dynamics can be more completely represented when multiple movement dimensions are considered simultaneously, especially in clinical conditions with partially overlapped motor manifestations such as IPD and VaP.

The results presented in Table \ref{tab:Results2variables} show the performance of the classifier when combining pairs of variables, both in the Off and On states. A consistent improvement is observed in each of the metrics when moving from the Off to the On state, suggesting that medication response contributes positively to the differentiation between IPD and VaP when using combinations of topological descriptors. In the Off state, most combinations just exceed 66\% accuracy, with several pairs below 60\%. Some combinations stand out, such as MaxTLSW + MinTC and MaxTLSW + MaxHC, reaching AUCs of 0.76 and 0.73, respectively, suggesting that certain variables already begin to show discriminatory potential even before the medication effect. When switching to the On state, a substantial improvement is evident. Several combinations exceed 75\% accuracy and AUC, such as MaxTESW + MinTC (AUC = 0.84, Accu = 83\%) and MaxTLSW + MinTC (AUC = 0.89, Accu = 79\%), indicating that these combinations manage to capture more stable differential patterns between IPD and VaP after levodopa administration. Also notable is the improvement in combinations that were initially weak in the Off state, such as MaxHC + StrikeAngle, which increased from an accuracy of 48\% to 76\% in the On state, suggesting greater stability of the motor pattern in IPD under medication. 

\begin{table}[H]
\caption{Classifier performance for two concatenated variables}
{\small
\centering\
\begin{tabular}{ll|cccc|cccc}
\toprule
 &  & \multicolumn{4}{c|}{Off State} & \multicolumn{4}{c}{On State} \\
\cmidrule(lr){3-6} \cmidrule(lr){7-10}
Variable 1 & Variable 2 & Accu & AUC &  Sens & Spec  & Accu & AUC &  Sens & Spec  \\
\midrule
LiftOffAngle & MaxTESW       & 0.66 & 0.64 & 0.80 & 0.50 & 0.66 & 0.73 & 0.73 & 0.57 \\
LiftOffAngle & MinTC        & 0.62 & 0.62 & 0.67 & 0.57 & 0.69 & 0.79 & 0.73 & 0.64 \\
LiftOffAngle & MaxTLSW       & 0.69 & 0.73 & 0.67 & 0.71 & 0.79 & 0.86 & 0.80 & 0.79 \\
LiftOffAngle & MaxHC        & 0.48 & 0.52 & 0.47 & 0.50 & 0.69 & 0.70 & 0.80 & 0.57 \\
LiftOffAngle & StrikeAngle  & 0.59 & 0.63 & 0.60 & 0.57 & 0.72 & 0.74 & 0.80 & 0.64 \\
MaxTESW & MinTC              & 0.66 & 0.64 & 0.73 & 0.57 & 0.83 & 0.84 & 0.87 & 0.79 \\
{\bf MaxTESW} & {\bf MaxTLSW} & {\bf 0.59} & {\bf 0.66} & {\bf 0.60} & {\bf 0.57} & {\bf 0.72} & {\bf 0.85} & {\bf 0.73} & {\bf 0.71} \\
MaxTESW & MaxHC              & 0.62 & 0.59 & 0.73 & 0.50 & 0.55 & 0.63 & 0.60 & 0.50 \\
MaxTESW & StrikeAngle        & 0.66 & 0.64 & 0.73 & 0.57 & 0.66 & 0.67 & 0.73 & 0.57 \\
{\bf MinTC}  & {\bf MaxTLSW} & {\bf 0.69} & {\bf 0.76} & {\bf 0.67} & {\bf 0.71} & {\bf 0.79} & {\bf 0.89} & {\bf 0.80} & {\bf 0.71} \\
{\bf MinTC}  & {\bf MaxHC}  & {\bf 0.66} & {\bf 0.63} & {\bf 0.67} & {\bf 0.64} & {\bf 0.76} & {\bf 0.82} & {\bf 0.87} & {\bf 0.64} \\
{\bf MinTC}  & {\bf StrikeAngle} & {\bf 0.66} & {\bf 0.63} & {\bf 0.67} & {\bf 0.64} & {\bf 0.69} & {\bf 0.83} & {\bf 0.73} & {\bf 0.64} \\
{\bf MaxTLSW} & {\bf MaxHC}  & {\bf 0.60} & {\bf 0.73} & {\bf 0.67} & {\bf 0.71} & {\bf 0.72} & {\bf 0.74} & {\bf 0.73} & {\bf 0.71} \\
{\bf MaxTLSW} & {\bf StrikeAngle}   & {\bf 0.66} & {\bf 0.68} & {\bf 0.67} & {\bf 0.64} & {\bf 0.79} & {\bf 0.84} & {\bf 0.87} & {\bf 0.71} \\
MaxHC  & StrikeAngle        & 0.48 & 0.53 & 0.60 & 0.36 & 0.76 & 0.69 & 0.80 & 0.71 \\
\bottomrule
\end{tabular}
\smallskip
  \begin{flushleft}
    \scriptsize{Accu: Accuracy; Sens: Sensitivity; Spec:Specificity; MaxHC: Maximum Heel Clearance; MaxTESW: Maximum Early Swing; MinTC: Minimum Toe Clearance; MaxTLSW: Maximum Toe Late Swing.}
  \end{flushleft}
}
\label{tab:Results2variables}
\end{table}

\begin{figure}[ht]
    \centering
    % Fila 1: MinTC + MaxTLSW
    \begin{subfigure}[b]{0.45\textwidth}
        \centering
        \includegraphics[scale=0.5]{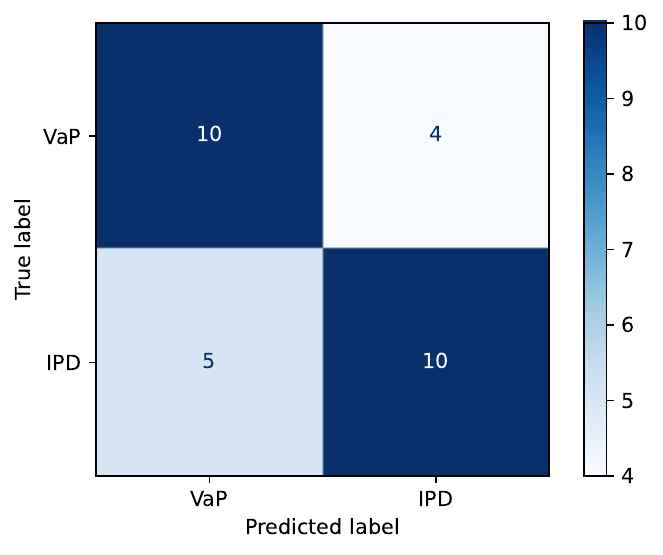}
        \caption{MinTC + MaxTLSW - Off State}
    \end{subfigure}
    \hfill
    \begin{subfigure}[b]{0.45\textwidth}
        \centering
        \includegraphics[scale=0.5]{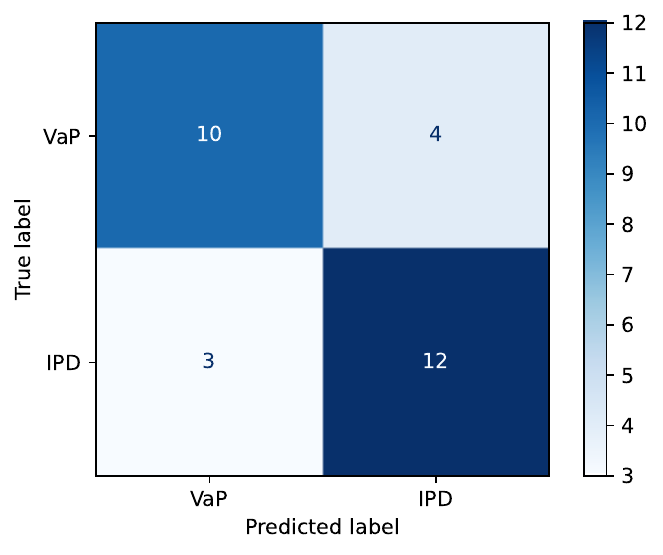}
        \caption{MinTC + MaxTLSW - On State}
    \end{subfigure}

    \vspace{0.5cm} % Espacio vertical entre filas

    % Fila 2: MaxTLSW + StrikeAngle
    \begin{subfigure}[b]{0.45\textwidth}
        \centering
        \includegraphics[scale=0.5]{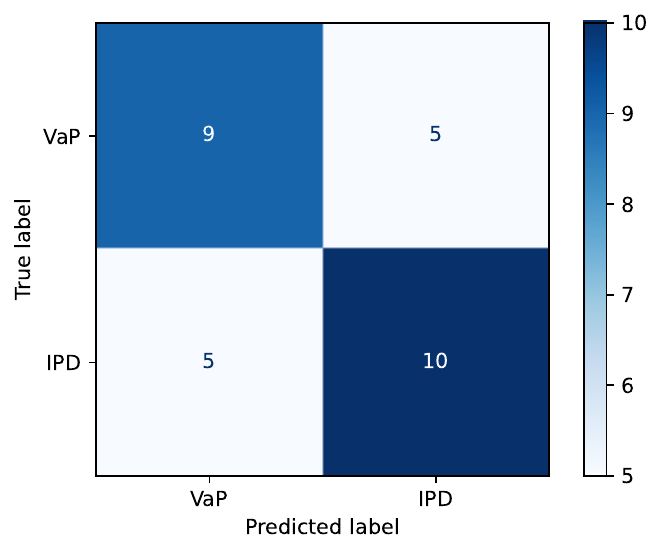}
        \caption{MaxTLSW + StrikeAngle - Off State}
    \end{subfigure}
    \hfill
    \begin{subfigure}[b]{0.45\textwidth}
        \centering
        \includegraphics[scale=0.5]{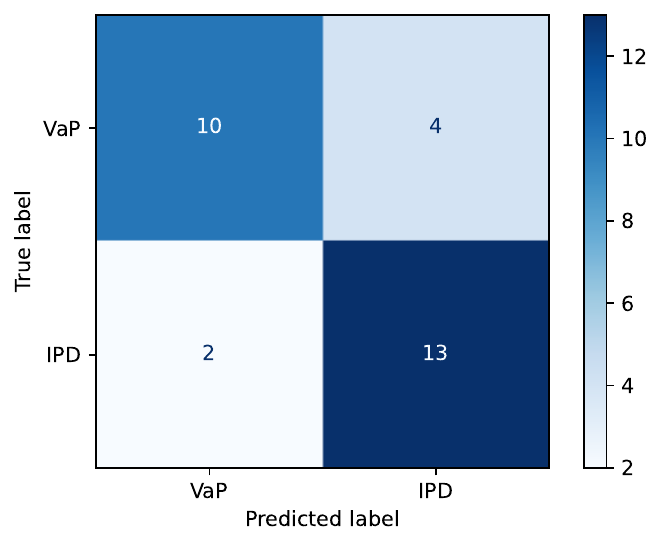}
        \caption{MaxTLSW + StrikeAngle - On State}
    \end{subfigure}

    \caption{Confusion matrices using combined variables in different medication states. Top row: MinTC + MaxTLSW; bottom row: MaxTLSW + StrikeAngle.}
    \label{fig:ConfMatrix_Combined}
\end{figure}

Figure \ref{fig:ConfMatrix_Combined} presents the confusion matrices for the combinations of variables MinTC and MaxTLSW (top row) and MaxTLSW and Strike Angle (bottom row), under both Off and On medication conditions. These results reinforce the observation that the On state, that is, after levodopa administration, allows the model to more clearly capture the differences between IPD and VaP. In Figure \ref{fig:ConfMatrix_Combined}a, corresponding to the Off state, the classifier correctly identified 10 patients with VaP and 10 with IPD, but significant errors were still observed (4 false positives for IPD and 5 false negatives), which is reflected in moderate sensitivity and specificity. However, in Figure \ref{fig:ConfMatrix_Combined}b, already in the On state, the true positives for IPD increase to 12 and the errors are reduced (only 3 false negatives), implying an improvement in both sensitivity and overall accuracy. For the combination of MaxTLSW and Strike Angle (bottom row) an even more significant improvement is observed. The Off state (Figure \ref{fig:ConfMatrix_Combined}c) show 5 false negatives and 5 false positives. However, in the On state (Figure \ref{fig:ConfMatrix_Combined}d), performance improves significantly, with 13 correct classifications for IPD and only 2 errors, implying high sensitivity (87\%) and improved specificity (71\%). This pattern confirms that medication has a clear impact on the dynamic structure of gait, which topological descriptors can effectively capture when the relevant variables are combined.

In terms of sensitivity and specificity, a greater symmetry is observed in the On state, with various combinations achieving balanced values above 70\% for both metrics. This reinforces the idea that the combination of variables allows the classifier to better capture topological differences distributed across multiple dimensions of movement. These results support the hypothesis that the integration of multiple sources of topological information by combining variables improves the performance of the model, especially in conditions where the effect of the medication amplifies the functional differences between IPD and VaP. Furthermore, it highlights that certain combinations (such as those involving MinTC, MaxTLSW and MaxTESW) are especially relevant to capture nonlinear dynamics useful in differential classification.

Following the previous analysis of 2-variable combinations, configurations with 3, 4, 5, and all gait variables were evaluated to determine their impact on the classification between IPD and VaP. The results of the best combination for each case are summarized in Table \ref{tab:Resultsmorevariables}, where it can be seen that the overall performance of the classifier improves in the On state compared to the Off state, confirming the stabilizing effect of the medication on gait and its positive influence on topological discrimination.

The best 3-variable combination was MinTC + MaxTLSW + MaxHC, achieving an accuracy of 76\% and an AUC of 0.89 in the On state, with balanced sensitivity and specificity (80\% and 71\%, respectively). This configuration not only improves the Off state but also highlights the key role of MinTC and MaxTLSW, which had previously demonstrated consistent performance. The addition of MaxHC enriches the information related to vertical foot motion, providing greater depth to the topological characterization. In the combination of 4 variables (LiftOffAngle + MinTC + MaxTLSW + Strike Angle), slightly higher performance is observed, with an accuracy of 79\% and an AUC of 0.82 in the On state, while maintaining high sensitivity and specificity values (80\% and 79\%). This result reflects that adding variables that capture the onset and offset of the stance phase allows the model to identify more detailed patterns of motor control. Interestingly, when using 5 variables, performance in the off state decreased (Accu = 0.62, AUC = 0.67), suggesting that including more information does not always entail a benefit in non-medication contexts. However, in the On state, classification improves considerably (Accu = 0.76, AUC = 0.87), again showing that the effect of levodopa facilitates the detection of topological differences between groups. Finally, the use of all available variables offers robust performance, especially in the On state, where 79\% accuracy, an AUC of 0.86, and high sensitivity (87\%) are achieved. These results suggest that a comprehensive multivariate view may be useful, although the gain compared to smaller combinations is marginal.
\begin{table}[H]
\centering
\caption{Classifier performance for the best combination}
{\small % \footnotesize
\begin{tabular}{c|cccc|cccc}
\toprule
Best result for & \multicolumn{4}{c|}{Off State} & \multicolumn{4}{c}{On State} \\
\cmidrule(lr){2-5} \cmidrule(lr){6-9}
combinaton of &  Accu & AUC &  Sens & Spec  & Accu & AUC &  Sens & Spec  \\
\midrule
{\tiny MaxHC + {\bf MinTC} + {\bf MaxTLSW} } & 0.76 & 0.74 & 0.80 & 0.71 & 0.83 & 0.89 & 0.80 & 0.71 \\
 & & & & & & & & \\
{\tiny LiftOffAngle + {\bf MinTC} + {\bf MaxTLSW} } & 0.72 & 0.73 & 0.73 & 0.71 & 0.79 & 0.82 & 0.80 & 0.79 \\
{\tiny + Strike Angle} & & & & & & & & \\

{\tiny LiftOffAngle + MaxHC + MaxTESW } & 0.62 & 0.67 & 0.60 & 0.64 & 0.76 & 0.87 & 0.80 & 0.71 \\
{\tiny + {\bf MinTC} + {\bf MaxTLSW}} & & & & & & & & \\

{\tiny LiftOffAngle + MaxHC + MaxTESW } & 0.69 & 0.64 & 0.67 & 0.71 & 0.79 & 0.86 & 0.87 & 0.71 \\
{\tiny + {\bf MinTC} + {\bf MaxTLSW} + Strike Angle} & & & & & & & & \\
\bottomrule
\end{tabular}
\smallskip
  \begin{flushleft}
    \scriptsize{Accu: Accuracy; Sens: Sensitivity; Spec:Specificity; MaxHC: Maximum Heel Clearance; MaxTESW: Maximum Early Swing; MinTC: Minimum Toe Clearance; MaxTLSW: Maximum Late Swing.}
  \end{flushleft}
}
\label{tab:Resultsmorevariables}
\end{table}
These findings reinforce that MinTC and MaxTLSW are the most influential variables in this classification task, since they appear in all the highest-performing combinations. Their relevance lies in the fact that they represent essential features of foot clearance and fine motor control during the swing phase, key aspects in differentiating between IPD and VaP. The use of topological descriptors derived from these variables, especially after levodopa administration, appears to be a promising strategy for differential diagnosis in complex clinical settings.

\section{Discussion}
This study demonstrates that the use of topological descriptors extracted through persistent homology, particularly Betti curves, provides an effective approach to characterizing nonlinear gait dynamics in patients with Parkinsonism with a smaller dataset. The ability of these tools to distinguish between IPD and VaP was improved under the effects of medication, demonstrating that the framework used is a complementary tool for differential diagnosis. These findings highlight the clinical relevance of foot clearance variables, the sensitivity of TDA to the effects of medication, and the added value this approach can offer compared to traditional methods in settings with phenotypic overlap.

\subsection{Foot Clearance Variables}
Foot clearance parameters revealed strong discriminative capacity in differentiating parkinsonian subtypes. Prior research showed that VaP patients tend to exhibit reduced clearance angles, while IPD patients demonstrate relatively preserved or exaggerated vertical foot motion under specific conditions \cite{ferreira2019, alcock2016, ogata2022foot}. Here, persistent homology captured these subtype-related signatures as global, nonlinear structure in the time series. To our knowledge, this is the first application of TDA to foot clearance time series for differential diagnosis, opening a practical route to extract clinically informative patterns beyond traditional statistics \cite{fernandes2021}.

\subsection{The effect of Levodopa}
Levodopa has a well-documented impact on gait in Parkinsonism, primarily influencing parameters such as gait speed, stride length, double support time, and foot clearance \cite{schaaf2017,ferreira2019,su2023deep}. These motor improvements reflect dopaminergic modulation of both spatial and temporal gait dynamics. However, the response to levodopa differs across parkinsonian subtypes. Patients with IPD typically exhibit more pronounced improvements, whereas those with VaP often show a blunted or inconsistent response \cite{ferreira2019}. This pharmacological asymmetry suggests that levodopa responsiveness may serve as a useful clinical marker for differential diagnosis.

\noindent  Our results suggest that topological descriptors derived from persistent homology are sensitive to these medication-induced changes. In line with previous findings using regression-based models \cite{fernandes2021}, classification accuracy improved significantly in the On state. Notably, the highest performance was achieved when Off and On state data were combined, highlighting the ability of topological features to capture both baseline and modulated gait dynamics.

\noindent  This suggests that the benefits of levodopa extend beyond simply increasing movement amplitude; they also include changes in the nonlinear structure and variability of gait—features that traditional linear models may fail to capture. The TDA framework thus offers a powerful, low-bias method to detect and monitor therapeutic response, with potential applications in both clinical assessment and longitudinal tracking of disease progression.

\subsection{Differential Diagnosis}
The results obtained reinforce the potential of the TDA-based framework as a complementary tool for the differential diagnosis of Parkinsonism. In contrast to traditional linear approaches, such as multiple linear regression (MLR) or structured normalization techniques \cite{fernandes2021,barrios2025}, TDA does not rely on assumptions about data distribution or the need for anthropometric corrections. While previous research has demonstrated the utility of TDA for differentiating neurodegenerative conditions \cite{yan2020classification,yan2022topological}, the present work extends these insights to the more clinically nuanced task of distinguishing IPD from VaP, a distinction complicated by overlapping phenotypes.

\noindent Evaluating the performance of topological descriptors in the classification task showed that Betti curves consistently outperformed the other descriptors (PL and SL) in all classification tasks. In each of the cases evaluated when classifying Parkinson's disease and Healthy control, BC showed AUCs greater than 0.99, similar to that reported by \cite{yan2020classification} who reached AUC of 0.9667, but using a time series with greater information and as a topological descriptor PL, which allows identifying the potential of each of the descriptors taking into account the nature and size of the time series used. Likewise, compared to the work of \cite{yan2022topological} where Freezing-of-Gait episodes were studied, where sensitivity and specificity reached 91.9\% and 87.6\% respectively using a fusion of topological descriptors, this study shows that even without fusion, BC vectors can effectively capture information from the gait time series, which reinforces its clinical applicability. For the more challenging IPD vs VaP task, combinations of foot clearance variables, particularly MinTC and MaxTLSW, were most informative. In the On state, the triplet MinTC + MaxTLSW + MaxHC reached AUC up to 0.89 with 83\% accuracy, competitive with \cite{fernandes2021}, which used a convolutional neural network (CNN) on gait time series and reported 82.33\% accuracy in the On state. Notably, our approach attains comparable performance while offering greater interpretability and reduced dependence on data normalization.

\subsection{Study limitations}
Although the results obtained are promising, this study has some limitations that should be considered. First, the sample size, although adequate for an exploratory analysis, is relatively small, which may limit the generalization of the findings to broader populations. This is especially relevant in the case of the classification between IPD and VaP, given the high degree of clinical variability within each subtype. Second, although a systematic methodology was applied to combine topological variables and multiple configurations were evaluated, the study focused exclusively on descriptors derived from persistence homology (BC, PL, and SL), without exploring other topological representations such as persistence images or persistence entropy that could enrich the analysis. 
Finally, although the effect of medication was addressed, it was assessed at the group level and did not consider intra-individual variability in response to levodopa, which could be relevant for personalized applications. Future studies with larger samples, longitudinal analyses, and hybrid approaches could overcome these limitations and further strengthen the utility of the TDA in real-world clinical settings.
\section{Conclusion}
This work evaluated the potential of TDA, particularly persistent homology, to characterize gait dynamics in patients with Parkinson's disease, using variables related to foot clearance. These variables, rarely used in the literature, proved to be highly informative in differentiating between parkinsonian subtypes. Through the generation of persistence diagrams and the subsequent extraction of topological descriptors, it was possible to capture nonlinear aspects of the movement that are not usually identifiable using traditional metrics. Integrating these descriptors into a Random Forest classifier allowed effective separation of patients from healthy controls and supported discrimination between idiopathic and vascular subtypes. Betti curves delivered the strongest overall performance across tasks, reflecting their ability to summarize the appearance and disappearance of connected components and loops ($H_0$ and $H_1$) and thus robustly represent the dynamics of gait.  The variables MinTC and MaxTLSW were consistently the most influential. Furthermore, the effect of medication was found to significantly improve the accuracy in distinguishing the model between IPD and VaP, suggesting that medication-induced changes can be detected sensitively using topological tools. This work expands the scope of previous research by proposing a complementary differential diagnosis framework, based on TDA and machine learning, that allows the identification of complex dynamic gait patterns with a higher level of detail. These findings open up new possibilities for the development of clinical support tools for the analysis and follow-up of patients with Parkinsonism.

\section*{Acknowledgements}
We thank the Foundation for Science and Technology (FCT) for the financial support provided through the doctoral scholarship with reference 2023.02242.BDANA (doi: \href{https://doi.org/10.54499/2023.02242.BDANA}{10.54499/2023.02242.BDANA}) and the support of Portuguese Funds through FCT within the Project UID/00013/2025 (doi: \href{https://doi.org/10.54499/UID/00013/2025}{10.54499/UID/00013/2025}).

%%%% References %%%%
\bibliographystyle{unsrt}
\bibliography{references}
\end{document}